\documentclass[conference]{IEEEtran}
\IEEEoverridecommandlockouts
\usepackage{cite}
\usepackage{amsmath,amssymb,amsfonts}
\usepackage{algorithmic}
\usepackage{graphicx}
\usepackage{textcomp}
\usepackage{xcolor}
\def\BibTeX{{\rm B\kern-.05em{\sc i\kern-.025em b}\kern-.08em
    T\kern-.1667em\lower.7ex\hbox{E}\kern-.125emX}}
\begin{document}

\title{Implicit Neural Representations for Speed-of-Sound Estimation in Ultrasound
}

\author{\IEEEauthorblockN{Michal Byra\IEEEauthorrefmark{1}\IEEEauthorrefmark{2},  Piotr Jarosik\IEEEauthorrefmark{2}\IEEEauthorrefmark{3}, Piotr Karwat\IEEEauthorrefmark{2}\IEEEauthorrefmark{3}, Ziemowit Klimonda\IEEEauthorrefmark{2}\IEEEauthorrefmark{3}, Marcin Lewandowski\IEEEauthorrefmark{3} }

\\

\IEEEauthorblockA{\IEEEauthorrefmark{2}Institute of Fundamental Technological Research, Polish Academy of Sciences, Warsaw, Poland}

\IEEEauthorblockA{\IEEEauthorrefmark{3}us4us Ltd., Warsaw, Poland}

\IEEEauthorblockA{\IEEEauthorrefmark{1}Corresponding author, e-mail: mbyra@ippt.pan.pl}
}

\maketitle

\begin{abstract}

Accurate estimation of the speed-of-sound (SoS) is important for ultrasound (US) image reconstruction techniques and tissue characterization. Various approaches have been proposed to calculate SoS, ranging from tomography-inspired algorithms like CUTE to convolutional networks, and more recently, physics-informed optimization frameworks based on differentiable beamforming. In this work, we utilize implicit neural representations (INRs) for SoS estimation in US. INRs are a type of neural network architecture that encodes continuous functions, such as images or physical quantities, through the weights of a network. Implicit networks may overcome the current limitations of SoS estimation techniques, which mainly arise from the use of non-adaptable and oversimplified physical models of tissue. Moreover, convolutional networks for SoS estimation, usually trained using simulated data, often fail when applied to real tissues due to out-of-distribution and data-shift issues. In contrast, implicit networks do not require extensive training datasets since each implicit network is optimized for an individual data case. This adaptability makes them suitable for processing US data collected from varied tissues and across different imaging protocols.

We evaluated the proposed SoS estimation method based on INRs using data collected from a tissue-mimicking phantom containing four cylindrical inclusions, with SoS values ranging from 1480 m/s to 1600 m/s. The inclusions were immersed in a material with an SoS value of 1540 m/s. In experiments, the proposed method achieved strong performance, clearly demonstrating the usefulness of implicit networks for quantitative US applications.





\end{abstract}


\section{Introduction}


Accurate estimation of the speed-of-sound (SoS) is crucial in ultrasound (US) imaging. SoS serves as a valuable quantitative parameter for tissue characterization, such as in the assessment of fatty liver disease \cite{telichko2022noninvasive}. More importantly, SoS is central to US image generation techniques, as it is required for performing high quality beamforming. However, US scanners typically set the SoS value to 1540 m/s, an average for soft tissues, for image generation tasks. This approximation degrades US image quality, ultimately affecting methods that process the reconstructed data \cite{byra2020adversarial}.

Various approaches have been proposed for SoS estimation, ranging from tomography-inspired algorithms like CUTE to convolutional networks and, more recently, physics-informed optimization frameworks based on differentiable beamforming~\cite{jaeger2015computed,karwat2023algorithm,simson2023differentiable,simson2024investigating}. However, the efficacy of current tomography-inspired techniques is limited due to the oversimplified physical models used in their derivation. Deep learning methods, primarily relying on encoder-decoder convolutional networks, have been designed to process raw US data for SoS estimation \cite{simson2024investigating}. However, the diversity of US imaging techniques and the variability of human anatomy complicate their practical application. Convolutional networks, pre-trained on simulated data, may fail when applied to real tissues due to out-of-distribution and data-shift issues. Additionally, convolutional networks may generate SoS parametric maps that are implausible from a physical perspective, as the physics is implicitly injected through the training data rather than a process that constrains outputs based on specific input RF data. Simson et al. utilized differentiable beamforming for US autofocusing (DBUA), in which the SoS estimation task is stated as an optimization problem \cite{simson2023differentiable}. This framework is based on differentiable beamforming, where the initial SoS value is pre-defined on a fixed grid of coordinates, and directly optimized with a loss function that associate the spatial SoS distribution with selected characteristic of the beamformed data, such as speckle brightness.   

\begin{figure*}[] \begin{center} \includegraphics[width=1\linewidth]{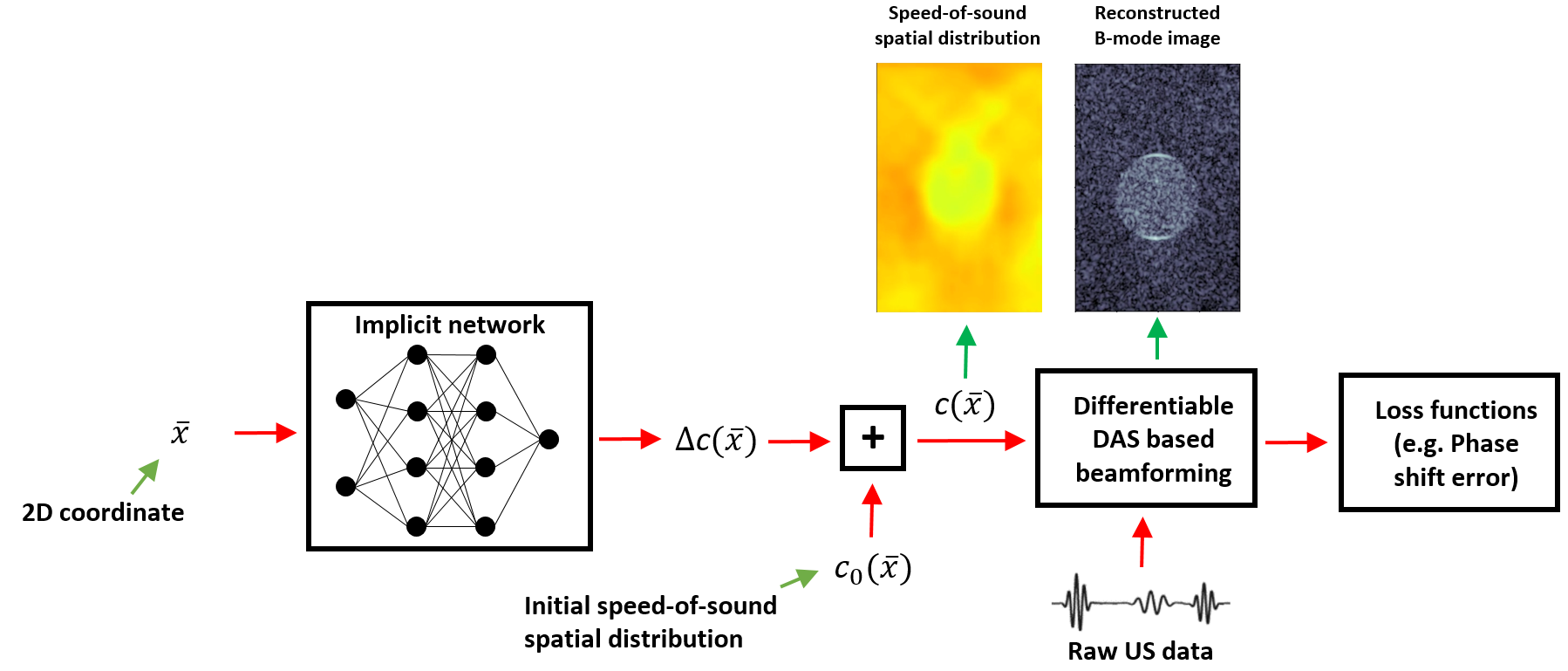} \end{center} \caption{A schematic of the proposed framework for speed-of-sound estimation in ultrasound. An implicit network is trained to produce speed-of-sound estimates, which are then used for image reconstruction through a differentiable delay-and-sum beamformer. The phase shift error is computed from the reconstructed data and backpropagated to the implicit network, updating its weights accordingly.} \label{f1} \end{figure*}

Implicit neural representations (INRs) are gaining momentum in biomedical image analysis. INRs are a type of neural network architecture that encodes continuous functions in the network's weights. Recently, these models have been applied to tasks such as cardiac segmentation \cite{stolt2023nisf}, brain image registration \cite{byra2023exploring}, and MRI image enhancement \cite{ye2023super}. Unlike traditional neural networks, INRs do not rely on discrete grid-based data, but instead represent information implicitly, allowing for more flexible and adaptable modeling of the target quantity \cite{sitzmann2020implicit}. INRs do not require extensive training datasets since each implicit network is optimized for a specific data case. This adaptability makes them well-suited for processing US data collected from a variety of tissues and across different imaging protocols. Compared to traditional quantitative US (QUS) techniques, implicit networks can utilize complex physics-related objective loss functions during optimization, enabling them to implicitly code nonlinear relationships that would be otherwise difficult to model using standard techniques.  

In this work, leveraging the differentiable beamforming framework, we employ implicit networks for SoS estimation in US. Our method extends the DBUA framework by incorporating the advantages of INRs, such as their ability to implicitly represent nonlinear mappings. Our initial results demonstrate that INRs can provide accurate and efficient computational modeling and parameter estimation in QUS.

\section{Methods}

\subsection{Implicit networks}

An INR is a multi-layer perceptron trained using input data coordinates to output the target quantity. In this work, we utilized the SIREN implicit network for SoS estimation. The SIREN model can be expressed using the following formula:

\begin{equation}
f_l(\bar{x})=
    \begin{cases}  
    \rho \left( W^{(l)} \bar{x} + b^{(l)}  \right), & l=1  \\
      \rho \left( W^{(l)} f_{l-1}(\bar{x}) + b^{(l)}  \right), & l \in \{2,...,L-1\}  \\
          W^{(l)} f_{l-1}(\bar{x}) + b^{(l)}, & l=L  \\
    \end{cases} 
\label{eq:eq1}
\end{equation}

\noindent where $\bar{x} \in [0,1]^2$ represents the input coordinates defined on a normalized 2D grid. The function $\rho(z)= \sin(\omega z)$ stands for the point-wise sine activation function, with the hyperparameter $\omega$ set to 30, following the original paper \cite{sitzmann2020implicit}. $W^{(l)}$ and $b^{(l)}$ correspond to the weight and bias of the $l$-th layer, respectively.

\subsection{Speed-of-Sound estimation}

In this study, we aim to estimate the spatial distribution of the speed-of-sound parameter in tissue using implicit networks. We formulate the problem as follows:

\begin{equation}
c(\bar{x}) = \Delta c(\bar{x}) + c_0(\bar{x}),
\label{eq:eq2}
\end{equation}

\noindent where $\Delta c(\bar{x})$ is represented by a coordinate-wise SIREN model, and $c_0(\bar{x})$ denotes the initial SoS spatial distribution (e.g., set to a constant value of 1540 m/s). In our approach, we train the implicit network to provide corrections to the initial estimate of the speed-of-sound distribution. Given the SoS distribution $c(\bar{x})$, we apply delay-and-sum (DAS) beamforming to compute the time-of-flight values and generate the B-mode image. As in the DBUA framework, the implicit network is optimized using the following loss function:

\begin{equation}
\mathcal{L}_{}(\bar{c}) = \mathcal{L}_{\text{pe}}(\bar{c}) + \alpha  \mathcal{L}_{\text{tv}}(\bar{c}) 
\label{eq:eq3}
\end{equation}

\noindent where $\mathcal{L}{\text{pe}}(\bar{c})$ is the phase shift error loss function, and $\mathcal{L}{\text{tv}}(\bar{c})$ represents the total variation regularization loss. The regularization weight $\alpha$ was set to 0.01. For a detailed description of the loss functions, we refer to the DBUA paper \cite{simson2023differentiable}.

\subsection{Training and Evaluation}

\begin{figure*}[]
	\begin{center}
		\includegraphics[width=0.9\linewidth]{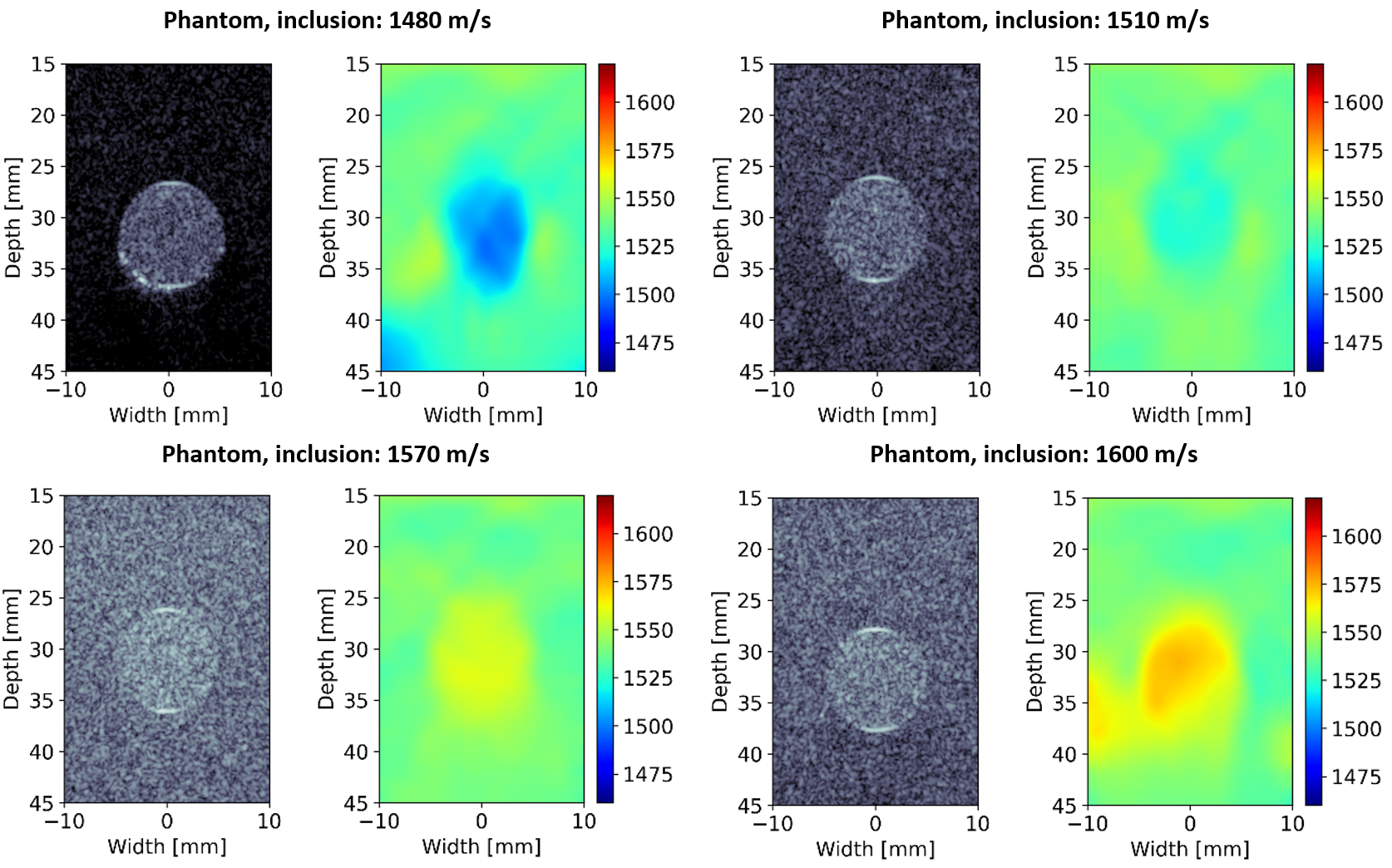}
	\end{center}
	\caption{Qualitative results obtained for the proposed method based on implicit networks. Reconstructed B-mode images and the speed-of-sound parametric maps were determined for a tissue-mimicking phantom containing inclusions with varying  speed-of-sound values.}
	\label{f2}
\end{figure*}

Experiments were performed using the us4R-lite US scanner (us4us, Poland) equipped with a 192-element linear probe SL1543 (Esaote, Italy) operating at a frequency of 5 MHz. The synthetic aperture technique was applied to acquire raw US data from a tissue-mimicking phantom (model 1438, Dansk Fantom Service, Denmark), containing cylindrical inclusions (diameter of 1 cm) with SoS values of 1480 m/s, 1510 m/s, 1570 m/s, and 1600 m/s, respectively. All inclusions were immersed in the material with the SoS value of 1540 m/s.

The proposed approach was compared with the DBUA method. The collected US data were pre-processed in the same way as in the DBUA technique to enable the use of the differentiable beamforming framework implemented in Python/JAX \cite{simson2023differentiable,jax2018github}. For estimation, we used the SIREN model with three hidden layers, each consisting of 128 units. Both the proposed method and the DBUA technique were optimized on a physical grid of 30 mm x 20 mm, corresponding to a 30x20 grid of 2D points used for training. Additionally, the output of the SIREN was multiplied by a factor of 100 to improve training, as the original weight initialization scheme favored outputs from a small interval. The networks were trained for 1000 epochs using Adam optimizer with learning rate set to 0.001 

To evaluate the investigated SoS estimation methods, we used the root mean squared error (RMSE) metric, with SoS values provided by the phantom manufacturer serving as the reference.

\section{Results}

Results obtained for the inclusion phantom are summarized in Table \ref{t1}. The proposed method, based on the implicit network, outperformed the DBUA technique in three out of four cases with respect to the RMSE metric. For the phantom containing the inclusion with an SoS value of 1600 m/s, our approach performed on par with DBUA. While the DBUA technique achieved similar scores across all four inclusions, our method clearly performed better for cases with lower SoS value differences between the inclusion and the background. The network struggled with accurately modeling high SoS variations. However, we believe this issue could be mitigated by using a nonlinear transformation of the target SoS values, which we leave for future work.

\begin{table}[]
\begin{center}

        \caption{RMSE scores were calculated for both the proposed method, based on implicit networks, and the DBUA technique.  The results were obtained using a tissue-mimicking phantom with inclusions with varying speed-of-sound values.}
        \label{t1}

\scalebox{1.1}{
\begin{tabular}{|l|c|c|}
\hline
    Phantom & Proposed (INRs) &  DBUA \\
                  \hline     
                  
    Inclusion, 1480 m/s & \textbf{15.4} & 17.6  \\  \hline     
    
    Inclusion, 1510 m/s & \textbf{8.0} & 15.7  \\  \hline     

    Inclusion, 1540 m/s  & \textbf{6.8} & 18.2 \\  \hline     
    
    Inclusion, 1600 m/s & \textbf{14.8} & \textbf{14.8}  \\  \hline  
    
\end{tabular}} 
\end{center}
\end{table}

\begin{figure}[]
	\begin{center}
		\includegraphics[width=1\linewidth]{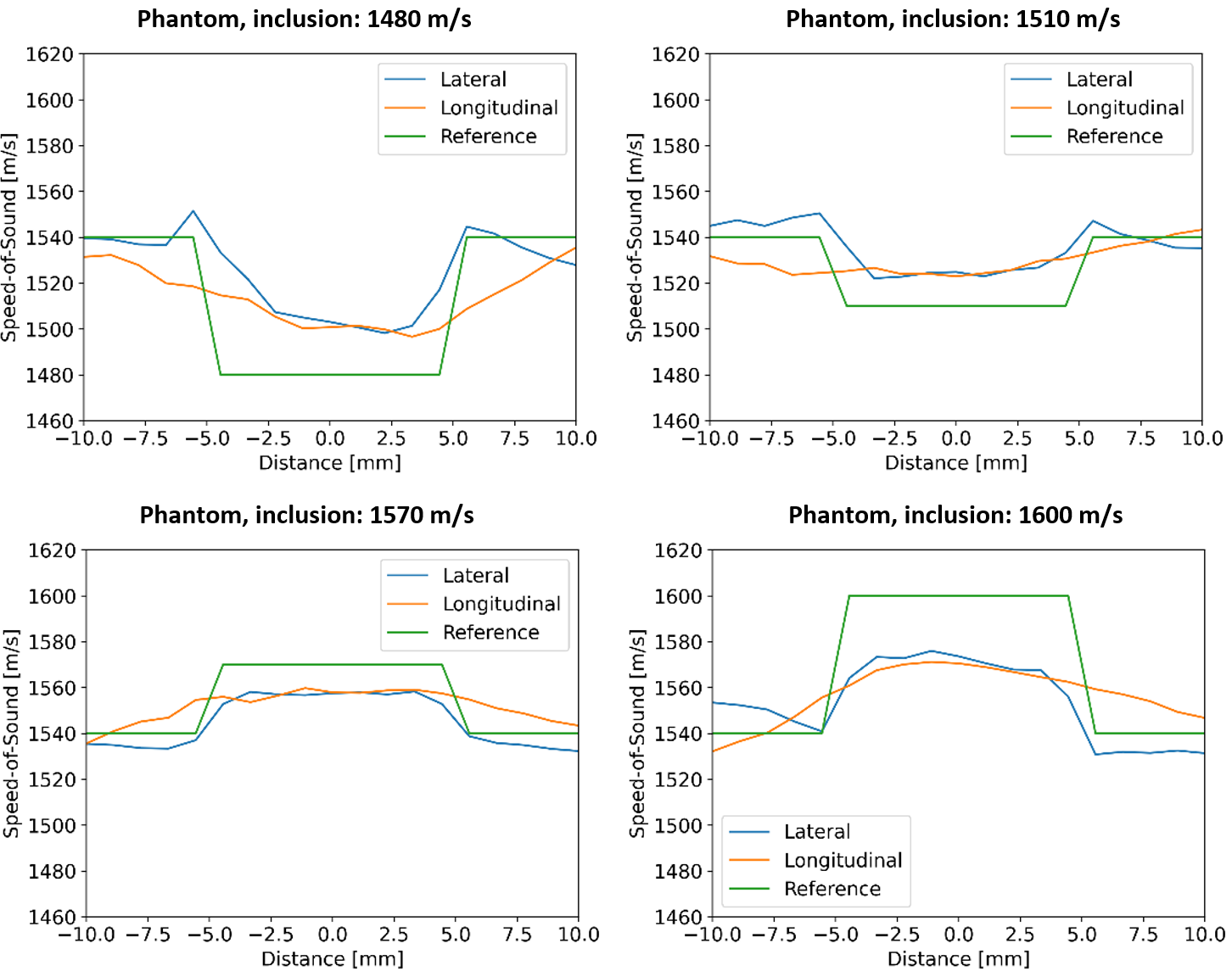}
	\end{center}
	\caption{The lateral and longitudinal cross-sections of the speed-of-sound parametric maps were analyzed with respect to the centers of the phantom inclusions. }
	\label{f3}
\end{figure}

Fig. \ref{f2} shows the reconstructed B-mode images and the SoS parametric maps determined using our method. The qualitative results demonstrate that the proposed method performed best for the inclusions with SoS values of 1510 m/s and 1570 m/s. For the two remaining cases, the method produced less circular spatial SoS distributions. Additionally, Fig. \ref{f3} illustrates the lateral and longitudinal cross-sections determined with respect to the centers of the phantom inclusions. Here, we observe that the implicit network tends to underestimate the target SoS values. Furthermore, the results obtained for the lateral dimension were better compared to the longitudinal axis. 

\section{Discussion}

In this paper, we demonstrated that implicit neural networks can be effectively used for SoS estimation in US. Our preliminary results indicate that implicit networks offer a promising framework for calculating quantitative US (QUS) parameters. Compared to standard QUS estimation techniques, physics-informed networks are less reliant on assumptions typically required to simplify modeling and formulate equations for SoS estimation \cite{jaeger2015computed}. Implicit networks present a nonlinear mapping, which is determined on a per-case basis through an optimization process driven by a specific loss function. Well-designed loss functions, inherently related to US physics, are therefore crucial for developing implicit networks for QUS parameter estimation. The method proposed in this work can be seen as a natural extension of the DBUA technique, where the SoS value was directly optimized on a predefined grid. In our approach, the use of an implicit network allows for learning a more representative mapping function for SoS estimation.

Our approach has several limitations. First, the training of the implicit network depends heavily on the employed loss functions. Adjusting the balance between the loss function components may lead to different outcomes or even cause divergence during training. However, this is a common issue with physics-informed networks \cite{cuomo2022scientific}. Similar to instance segmentation in neural networks, special adjustments may be needed to fine-tune the loss function to accurately estimate SoS values for small objects \cite{rachmadi2024new}. Second, we did not evaluate the proposed method on US data collected from humans, leaving it uncertain whether the approach can be applied to human tissues with complex anatomy. Third, while our implementation followed the DBUA framework, better results might be achievable through careful optimization of hyperparameters.

In the future, we plan to explore novel loss functions for training implicit networks. For example, we intend to design loss functions that jointly address US image reconstruction and the estimation of other QUS parameters. We also aim to apply INRs to decompose raw US data into components representing different physical properties, similar to recent work in texture separation and image registration in microscopy \cite{byra2023implicit}. Additionally, we will extend our method to 3D US imaging and consider applications like temperature monitoring \cite{byra2022unsupervised}. The application of INRs in US represents a promising direction for further research.

\section*{Acknowledgement}

M. Byra has no conflicts of interest to disclose. P. Jarosik, P. Karwat. Z. Klimonda and M. Lewandowski are employed at us4us Ltd., the manufacturer of the ultrasound scanner used to collect data for this study. 

\bibliographystyle{IEEEtran}
\bibliography{IEEEabrv,references}

\end{document}